\documentclass{article} 
\usepackage{amsmath, amsthm, amssymb, graphicx, url, algorithm, hyperref,url}
\usepackage[authoryear]{natbib}

\usepackage{algorithmic}
\let\emptyset\varnothing
\let\epsilon\varepsilon
\let\phi\varphi

\def\G{{\mathcal G}}
\def\L{{\mathcal L}}

\def\O{{\mathcal O}}

\def\RR{\mathcal R}
\def\S{\mathcal S}
\def\U{{\mathcal U}}
\def\I{\mathcal I}
\def\x{{\boldsymbol{x}}}

\def \kk{\kappa}

\def\R{\mathbb R}
\def\N{\mathbb N}

\def\B{\mathcal B}
\def\X{\mathcal X}

\def\-as{\text{-a.s.}}
\def\argmax{\operatorname{argmax}}
\def\argmin{\operatorname{argmin}}

\newcommand{\llft}[1]{b_i}
\newcommand{\rrt}[1]{b_{i+1}}
\newcommand{\mingap}{\lambda_{\min}}
\newcommand{\gap}{\lambda}
\newcommand{\sort}{\text{\textbf{sort}}}

\newcommand{\argmaxdisp}[1]{\underset{#1}{\argmax~}}
\newcommand{\argmindisp}[1]{\underset{#1}{\argmin~}}
\newcommand{\maxdisp}[1]{\underset{#1}{\max~}}

\newcommand{\numdist}{r}
\newcommand{\numkappa}{m}

\title
{A consistent clustering-based approach to estimating the number of change-points in highly dependent time-series}

\author{Azadeh Khaleghi \\ azadeh.khaleghi@inria.fr \and
Daniil Ryabko \\ daniil.ryabko@inria.fr}
\date {INRIA, Lille Nord-Europe}

\begin{document}
\newtheorem{thm}{thm}
\newtheorem{lem}{lem}
\newtheorem{defn}{defn}
\newtheorem{prop}{Proposition}

\maketitle

\begin{abstract}
The problem of change-point estimation is considered
under a general framework where the 
data are generated by unknown stationary ergodic process distributions.
In this context, the consistent estimation of the number
of change-points is provably impossible.
However, it is shown that a consistent clustering method 
may be used to estimate the number of change points, 
under the additional constraint that the correct number of 
process distributions that generate the data is provided.
This additional parameter has a natural interpretation 
in many real-world applications. 
An algorithm is proposed that estimates 
the number of change-points and locates the changes. 
The proposed  algorithm is shown to be asymptotically consistent;
its empirical evaluations are provided. 
\end{abstract}
\section{Introduction}
Change-point estimation is a classical problem 
in statistics and machine  learning, with applications in a 
broad range of domains, such as market analysis, bioinformatics, audio and video segmentation, 
fraud detection, only to name a few. 
The change-point problem may be described as follows.
A sequence 
$\x:=
X_1, 
\dots,
X_n
$ 
is  composed of some (unknown) number $\kappa+1$ of non-overlapping segments. 
Each segment is generated by one of $\numdist$ (unknown) stochastic process distributions. 
The process distributions that generate every pair of consecutive segments 
are different.  The index where one segment ends and another starts is called a {\em change point}.
The change-points are unknown, and the objective is to estimate them given $\x$.

In this work we consider the change-point problem for highly dependent data, making as little assumptions as possible
on how the data are generated. 
In particular, the distributions that generate the data are unknown and can be arbitrary; 
the only assumption is that they are stationary ergodic. This means that  we make no such assumptions as independence, finite memory or mixing. Moreover, 
we do not require the finite-dimensional marginals of any fixed size before and after the change points to be different.

However, with no further assumptions or additional information, 
the estimation of the number of change-points 
is impossible even in the weakest asymptotic sense. 
Indeed, as shown by \cite{Ryabko:10discr},
it is  impossible to distinguish 
even between the cases of 0 and~1 change-point in this setting, even for binary sequences. 
As an alternative to imposing stronger assumptions on the distributions that would allow for the estimation 
of the number of change points,
 we 
assume that 
the {\em correct number $r$ of the process distributions}
that generate $\x$ is provided as a parameter. 

This formulation is motivated by applications. 
Indeed, 
the assumption that the time-series data are highly dependent 
complies well with most real-world scenarios. 
Moreover, in many applications the number $r$ of distributions is a natural 
parameter of the problem. For instance, the case of just $r=2$ distributions 
can be interpreted as normal versus abnormal behavior; one can
imagine a sequence with many change-points in this scenario. 
Another application concerns the problem of author attribution in a given text 
written collaboratively by a known number $\numdist$ of authors. 
In speech segmentation $\numdist$ may be the total number of speakers. 
In video surveillance as well as in fraud detection, 
the change may refer to the point where normal activity becomes abnormal ($\numdist$=2). 
The identification of  coding versus non-coding regions in genomic data is 
yet another potential application. 
In other words, in many real-world applications 
the number $\numdist$ of process distributions 
comes with a natural interpretation.

\textbf{Main Results.}
We propose a nonparametric algorithm to
 estimate the number of change points and 
to locate the changes 
in time-series data. 
We demonstrate both theoretically and experimentally 
that our algorithm is asymptotically consistent
in the general framework described. 
A key observation we make is that 
given the total number $\numdist$ of process distributions, estimating the number of 
change-points is possible via a consistent time-series clustering method.
We use a so-called list-estimator to generate 
an exhaustive list of change-point candidates.
This induces a partitioning of the sequence into consecutive segments.
We then apply a simple clustering algorithm to group these segments into $\numdist$ clusters.  
The clustering procedure uses farthest-point initialization to designate $r$ cluster centers, 
and then assigns each remaining point to the nearest center.
To measure the distance between the segments, 
empirical estimates of the so-called distributional distance \cite{Gray:88} are used \citep[cf.][]{Ryabko:10clust}. 
In each cluster, we identify the change-point candidate
that joins a pair of consecutive segments as {\em redundant}. 
Finally, we remove the redundant estimates from the list 
and provide the remaining estimates as output.
The consistency of the proposed method can be established
using any list-estimator 
that is consistent under the considered framework,
in combination with the time-series clustering algorithm mentioned above.
An example of a consistent list-estimator is provided by \cite{khaleghi:12mce}.
Thus, the proposed method establishes a new link between two classical  unsupervised learning problems: clustering and change-point analysis,
potentially bringing a new insight to both communities.

\textbf{Related Work.}
In a typical formulation of the change-point problem 
the samples within each segment are assumed to be generated i.i.d, 
the distributions have known forms and the change is in the mean. 
In more general nonparametric settings, 
the form of the change and/or the nature of 
dependence are usually restricted. For example. the process distributions are assumed to
be strongly mixing \citep{brodsky:93,basseville:93,Giraitis:95,HarizWylieZhang2007,Carlstein:93}, 
and the finite-dimensional marginals are almost exclusively assumed to be different. 
The problem of estimating the number of change-points 
is nontrivial, even under these more restrictive assumptions. 
In such settings, this problem is usually
addressed with penalized criteria; see, for example,  \citep{Lebarbier2005717,Lavielle20051501}.
Such criteria necessarily rely on additional parameters, and the resulting 
number of change-points depends on these parameters. 
Note that the algorithm proposed in this work also requires an input parameter: the 
number $\numdist$ of distributions. 
However, this parameter has a natural interpretation 
in many  real-world applications as discussed above.

For the general framework considered in this work,  
the particular case of a known number $\kappa$ of change points has been considered in \citep{Ryabko:103s} ($\kappa$=1) and \citep{arxiv12} ($\kappa>1$).
However, if the  number $\kk$ of change-points provided to the algorithm is incorrect, 
the behavior of these algorithms can be arbitrarily bad.
An intermediate solution for the case of unknown $\kappa$ in this general setting 
is given by \cite{khaleghi:12mce} where  
a  list estimator is proposed: a (sorted) list of possibly more than $\kappa$ candidate estimates is produced
whose first $\kappa$ elements are consistent estimates of the change-points. 
The algorithms in these works, as well as in the present paper, are based on empirical estimates of distributional distance, which turns
out to be a rather versatile tool for studying stationary ergodic time series.  

\textbf{Organization.} 
In Section~\ref{sec:pre} we introduce some preliminary notation and definitions. 
In Section~\ref{sec:protocol} we formalize the problem. 
In Section~\ref{sec:theoretical}  we 
present our algorithm and give an informal description and 
in Section~\ref{sec:proof} we prove the main consistency result. 
In Section~\ref{sec:exp} we present some experimental results and finally
in Section~\ref{sec:conc} we provide our conclusions.
 
\section{Preliminaries}\label{sec:pre}
Let  $\X$ be a measurable space (the domain); in this work we let $\X=\mathbb R$ 
but extensions to more general spaces are straightforward.
For a sequence $X_1,\dots,X_n$ we use the abbreviation $X_{1..n}$.
Consider the Borel $\sigma$-algebra $\B$ on $\X^\infty$ generated by the cylinders $\{B\times \X^\infty: B\in B^{m,l}, m,l\in\N\}$,  
where the sets $B^{m,l}, m,l \in \N$ are obtained via the partitioning of $\X^m$ into  cubes  
of dimension $m$ and volume $2^{-ml}$ (starting at the origin). Let also
 $B^m:=\cup_{l\in\N}B^{m,l}$. 
Process distributions are probability measures 
on the space $(\X^\infty,\B)$.
For  $\x = X_{1..n}\in \X^n$ and $B\in B^m$ let $\nu(\x,B)$ 
denote the {\em frequency} with which $\x$ falls in~$B$, i.e. 
\begin{equation}\label{eq:nu}
\nu(\x,B):=  {{\frac{\mathbb I\{n \geq m \}}{n-m+1}}}  \sum_{i=1}^{n-m+1} \mathbb I \{ X_{i..i+m-1} \in B \}
\end{equation}
A process $\rho$ is {\em stationary}
if for any $i,j\in 1..n$ and $B \in B^m,~m \in \N$, 
we have $\rho(X_{1..j} \in B)=\rho(X_{i..i+j-1} \in B).$
A stationary process $\rho$ is called {\em  ergodic} if for all $B\in\mathcal B$ 
with probability~1 we have 
$\lim_{n\rightarrow\infty}\nu(X_{1..n},B) = \rho(B).$ 
\begin{defn}[Distributional Distance]
The  distributional distance between a pair of process distributions
$\rho_1,\rho_2$ is defined as follows~\citep[see ][]{Gray:88}.
$$
d(\rho_1,\rho_2)=\sum_{m,l=1}^\infty w_m w_l \sum_{B\in B^{m,l}} |\rho_1(B)-\rho_2(B)|,
$$
where we set  $w_j:=1/k(k+1)$, 
 but any summable sequence of positive weights may be used.
\end{defn}
In words, this involves partitioning the sets $\X^m$, $m\in\N$ into cubes of 
 decreasing volume (indexed by $l$) and then taking a sum over 
 the differences in probabilities of all the cubes in these partitions. 
The differences in probabilities are weighted: smaller
weights are given to larger $m$ and finer partitions.
We use {\em empirical estimates} of this distance defined as follows.
\begin{defn}[Empirical estimates of $d(\cdot,\cdot)$]
The empirical estimate of the distributional distance between a 
sequence $\x=X_{1..n} \in \X^n, n\in \N$ and a process distribution $\rho$ is given by
\begin{equation}\label{eq:emd1}
 \hat d(\x,\rho):=\sum_{m,l=1}^\infty w_{m,l} \sum_{B\in B^{m,l}} |\nu(\x,B)- \rho(B)| 
\end{equation}
and that between a pair of sequences 
$\x_i \in \X^{n_i}~n_i \in \N,~i=1,2$. 
is defined as
\begin{equation}\label{eq:emd2}
 \hat d(\x_1,\x_2):=\vspace{-0.2cm}\sum_{m,l=1}^\infty w_{m,l}\hspace{-0.2cm} \sum_{B\in B^{m,l}} \hspace{-0.15cm}|\nu(\x_1,B)-\nu(\x_2,B)|
\end{equation}
\end{defn}
While the calculation of $\hat d(\cdot,\cdot)$ 
involves infinite summations 
it is fully tractable. \\
\textbf{Remark~1~(Calculating $\hat{d}(\cdot,\cdot)$)}
Consider a pair of sequences $\x_i:=X^i_1, \dots, X_{n_i}  \in \X^{n_i}$ with $n_i \in \N,~i=1,2$.  
Let $s_{\min}$ correspond to the partition where each cell $B \in \B$ contains at most one point i.e. 
$$s_{\min}:=\min_{\substack {u,v \in 1,2\\ i,j \in 1..\min\{n_1,n_2\}\\ X_i^u \ne X_j^v}}|X_i^u-X_j^v|$$
Indeed in~\eqref{eq:emd1} all summands corresponding to $m>\max_{i=1,2} n_i$ equal 0;
moreover, all summands corresponding to $l>s_{\min}$ are equal. 
Thus as shown by \cite{Ryabko:10clust} even the most naive implementation of $\hat{d}(\x_1,\x_2)$ has computational complexity
$\O(n^2\log n\log s_{\min})$ 
which may be further optimized to $\O(n \operatorname{polylog} n)$,
see \citep{khaleghi:12mce,Ryabko:10clust,khaleghi:12}.
\section{Problem Formulation}\label{sec:protocol}
We formalize the problem as follows.
The sequence $\x:=X_1,\dots,X_n \in \X^{n},~n\in \N$ is formed as
the concatenation of some {\em unknown number} $\kappa+1$ of 
sequences
$$
X_{1..n\theta_1},
X_{n\theta_1+1..n\theta_2}, \dots,
X_{n\theta_{\kappa}+1..n}
$$
where $\theta_k \in (0,1),~k=1..\kappa$. 
Each of the sequences $\x_k:=X_{n\theta_{k-1}+1..n\theta_k},~k=1..\kappa+1,
~\theta_0:=0,~\theta_{\kappa+1}:=1$ is generated 
by one out of $\numdist \leq \kappa+1$ {\em unknown stationary ergodic} process distributions $\rho_1,\dots,\rho_{\numdist}$. 
Thus, there exists a ground-truth partitioning 
\begin{equation}\label{eq:gt}
\{\G_1,\dots,\G_{\numdist}\}
\end{equation} of the set $\{1..\kappa+1\}$
into $\numdist$ disjoint subsets where for every $k=1..\kappa+1$ and $\numdist '=1..\numdist$ we have
$k \in \G_{\numdist '}$ if and only if $\x_k$ is generated by $\rho_{\numdist '}$. 
The parameters $\theta_k,~k=1..\kappa$ are called {\em change-points} 
since the indices $n\theta_k,~k=1..\kappa$ separate consecutive segments $\x_k,\x_{k+1}$ 
generated by {\em different} process distributions.
The change-points are unknown, and our goal is to estimate them given the sequence $\x$. 
The process distributions $\rho_1,\dots,\rho_{\numdist}$ are completely unknown
and may even be dependent. Moreover, the
means, variances, or more generally, the finite-dimensional
marginal distributions of any fixed size before and after the
change-points are not required to be different. We consider the most general
scenario where the process distributions are different.
Let the minimum separation of the change-points be defined as
\begin{equation}\label{defn:thetamin}
\mingap:=\min_{k=1..\kappa+1} \theta_k-\theta_{k-1} .
\end{equation}
Since the consistency properties we are after
are asymptotic in~$n$, we require that $\mingap>0$.
This is because if the length of one of the sequences is constant or 
sub-linear in $n$ then asymptotic consistency is impossible in this setting.
Note, however, that we do not make any assumptions on the distance between the process distributions (e.g., the distributional distance): they may be arbitrarily close. 

Since it is provably impossible \citep{Ryabko:10discr}  
to distinguish between the case of one and zero change-points
in this general framework, the number $\kappa$ of change-points cannot be estimated with no further information.
Instead of making additional assumptions on the nature of the distributions generating the data, we assume that the total number $\numdist$ of distributions is provided (while the number $\kappa$ of change-points remains unknown).

Thus, the {\em problem formulation} we consider is as follows: 
given a sequence $\x$, a lower-bound on the minimum separation of the change points $\lambda$,  and the 
total number of distributions $r$, it is required to find the number of changes $\kappa$ and estimate the change points $\pi_1,\dots,\pi_\kappa$. 
A change-point estimator is a function that takes a sequence $\x \in \X^n,~n \in \N$ to 
produce a number $\hat\kappa$ (estimated number of change points) 
and a set $\{\hat{\theta}_1(n),\dots,\hat{\theta}_{\kappa}(n)\} \subset (0,1)^{\hat\kappa}$
of  estimated change points.  
It is asymptotically consistent if  with probability $1$ we have $\hat\kappa=\kappa$ from some $n$ on and
\begin{equation*}
\lim_{n \rightarrow \infty} \sup_{k=1..\kappa}|\hat{\theta}_k(n)-\theta_k|=0.
\end{equation*}
The algorithm we propose relies on a so-called {\em  list-estimator},
which is a procedure that, given $\x$ and $\lambda$, outputs a (long, exhaustive)  list
of change point estimates, without attempting to estimate the number of changes.  
More precisely,  we have the following definition. 
\begin{defn}[List-estimator]\label{defn:gen} 
A list-estimator $\Upsilon$ is a function that, given a sequence $\x \in \X^n$ 
and a number $\gap \in(0,1)$, 
produces a set $\Upsilon(\x,\gap) \in \bigcup_{i \in \N}(0,1)^i$ of  some $\numkappa \in \N$ estimates
 $\Upsilon(\x,\gap):=\{\hat{\theta}_1(n),\dots,\hat{\theta}_{\numkappa}(n)\}$, that are at least $\lambda$ apart: 
\begin{equation*} 
\inf_{i \neq j \in 0..\numkappa+1}|\hat{\theta}_i(n)-\hat{\theta}_j(n)| \geq \gap 
\end{equation*}
where $ \hat{\theta}_0(n):=0,~\hat{\theta}_{\numkappa+1}(n):=1$. 

Let $\x$ have change-points at least $\mingap$ 
apart for some $\mingap \in (0,1)$. 
A  list-estimator $\Upsilon$ is said to be consistent if for every $\gap \in (0,\mingap)$ 
there is a subset
$\{\hat{\theta}_{\mu_1}(n),\dots,\hat{\theta}_{\mu_{\kappa}}(n)\}$ of $\Upsilon(\x,\gap)$
for some $\mu_i \in 1..m,~i=1..\kappa $ such that
with probability one we have
\begin{equation*}
\lim_{n\rightarrow \infty}\sup_{k=1..\kappa}|\hat{\theta}_{\mu_k}(n)-\theta_k| =0.
\end{equation*}
\end{defn}
An example of a consistent list-estimator is provided in \citep{khaleghi:12mce}.
In particular we use the following statement.
\begin{prop}[\cite{khaleghi:12mce}]\label{thm0}
There exists a consistent list-estimator~$\Upsilon$.
\end{prop}
\section{Main Result}\label{sec:theoretical}
 \begin{algorithm}[h]
\caption{  {Clustering-Based Change-Point (CluBChaPo) Estimator}} \label{alg:main}
\begin{algorithmic}
{  \STATE {{\small \bfseries input:} $\x \in \X^n$, $\gap \in (0,\mingap]$, Number  $\numdist$ of process distributions}
\STATE \textbf{{\small 1. Obtain an initial (sorted)
set of change-point candidates using a consistent list-estimator $\Upsilon$ (see Definition~\ref{defn:gen}):} }
\vspace{-0.3cm}
\begin{align*}
&\Psi \gets \Upsilon(\x,\gap)~\text{and let}~\numkappa\gets |\Psi|\\
&\{\psi_i:i=1..\numkappa\} \gets \sort(\{n\hat{\theta}:\hat{\theta} \in \Psi\}),~\text{so that}~i< j \Leftrightarrow \psi_i < \psi_j,~i,j \in 1..\numkappa.
\end{align*}
\vspace{-0.7cm}
\STATE {\textbf {{\small 2. Generate a set $\S$ of consecutive segments:}}}
\vspace{-0.3cm}
\begin{equation}\label{alg:main:eq:segs}
\S \gets \{\widetilde{\x}_i:=X_{\psi_{i-1}+1..\psi_i}: i=1..\numkappa+1,~\psi_0:=0,~\psi_{\numkappa+1}:=n\}
\end{equation}
\STATE\vspace{-0.8cm}
 {\textbf{{\small 3. Partition $\S$ into $\numdist$ clusters:}}}
\STATE{~~~~Initialize $\numdist$ farthest segments as cluster centers:
\vspace{-0.4cm}
\begin{equation}\label{alg:cj}c_1 \gets 1,~c_j \gets \argmax_{i=1..\numkappa} \min_{i'=1}^{j-1}\hat{d}(\widetilde{\x}_i,\widetilde{\x}_{c_{i'}}),~j=2..\numdist \end{equation}}
\vspace{-0.6cm}
\STATE{~~~~Assign every segment to a cluster:}
\vspace{-0.3cm}
\begin{equation*}T(\widetilde{\x}_{i}) \gets \argmin_{j=1..\numdist} \hat{d}(\widetilde{\x}_i,\widetilde{\x}_{c_j}),~i=1..\numkappa\end{equation*}
\STATE \vspace{-0.8cm}
 \textbf{{\small 4. Eliminate redundant estimates:}}\vspace{-0.3cm}
\renewcommand{\labelitemi}{~}
\begin{itemize}
\STATE{$\mathcal C \gets \{1..\numkappa\}$}\vspace{-0.4cm}
\FOR{$i=1..\numkappa$}
\IF{$T(\widetilde{\x}_i)=T(\widetilde{\x}_{i+1})$}
\STATE $\mathcal C \gets  \mathcal C \setminus\{i\}$
\ENDIF
\vspace{-0.3cm}
\ENDFOR
\end{itemize}
\vspace{-0.3cm}
\STATE{$\hat{\kappa} \gets |\mathcal C|$}
\STATE {\bfseries return: $\hat{\kappa}$,~$\{\hat{\theta}_i:=\frac{1}{n}\psi_i:i\in\mathcal C\}$}}
\end{algorithmic}
\end{algorithm}
In this section we introduce an asymptotically consistent algorithm for estimating the number of change points and locating the changes. 
\begin{thm}\label{thm1}
Let  $\x:=X_{1..n} \in \X^n,~n\in\N$ be a sequence with change-points at least $\mingap$ apart, for some $\mingap \in (0,1)$. 
Let $\numdist$ denote the total number of process distributions generating $\x$.
Then CluBChaPo$(\x,\gap,\numdist)$ is asymptotically consistent for all $\gap \in(0,\mingap]$.
\end{thm}
{ The proof of Theorem~\ref{thm1} is deferred to Section \ref{sec:proof}; here we provide an intuitive explanation of how the algorithm works and why it is consistent.}\\
The algorithm works as follows. 
First, a (consistent) list-estimator is used to obtain an initial set of change-point candidates. 
The candidates are sorted in increasing order to produce a set $\S$ 
of consecutive non-overlapping segments of $\x$. 
The set $\S$ is then partitioned into $\numdist$ clusters. 
In each cluster, the change-point candidate
that joins a pair of consecutive segments of $\x$ is identified as {\em redundant} 
and is removed from the list. 
Once all of the redundant candidates are removed, 
the algorithm outputs the remaining change-point candidates. 
Next we give an intuitive explanation as to why the algorithm works. 

Since the list estimator $\Upsilon$ is consistent,  
from some $n$ on an initial set of possibly more than $\kappa$ 
change-points are generated that is guaranteed to have a subset of size
$\kappa$ whose elements are arbitrarily close to the true change-points. 
Therefore, from some $n$ on the largest portion of each segment in 
$\S$ is generated by a single process distribution.
Since the initial change-point candidates 
are at least $n\gap$ apart,
the segments in $\S$ have lengths linear in $n$. 
Thus, we can show that from some $n$ on the 
distance between a pair of segments in $\S$ converges 
to $0$ if and only if the same process distribution generates most of the two segments. 
Given the total number of process distributions, 
from some $n$ on the clustering algorithm groups together those and only those segments in $\S$ 
that are generated by the same process distribution. 
This lets the algorithm identify and remove the redundant candidates. 
By the consistency of $\Upsilon$ the remaining estimates converge to the true change-points.

As an example of a consistent 
list-estimator the method proposed by \cite{khaleghi:12mce} may be used. 
This algorithm   
outputs a list of estimates whose first 
$\kappa$ elements converge to the true change-points, provided that
the parameter $\gap$ satisfies $\gap \in (0,\mingap]$.  
Since $\kappa$ is unknown, all we can use
here is that the correct change-point estimates are somewhere in the list. 
 In general the algorithm may use any list-estimator that is
 consistent (in the sense of Definition~\ref{defn:gen}) for stationary ergodic 
 time series.
In the proposed algorithm the following  consistent clustering procedure is used.
First,  a total of $\numdist$ cluster centers are obtained as follows.
The first segment $\x_1$ is the first cluster center. 
Through an iteration on $j=2..\numdist$ a segment is 
chosen as a cluster center if it has the highest minimum distance from 
the previously chosen cluster centers. 
Once the cluster centers are specified, the remaining 
segments are assigned to the closest cluster. \\
\textbf{Remark~2~(Computational Complexity)} 
In this implementation, an initial set of $\gap^{-1}$ change-point candidates is obtained by the algorithm of \cite{khaleghi:12mce}
which as shown by the authors has complexity $\O(n^2\operatorname{polylog}n)$. 
It is easy to see that the clustering 
procedure requires $\numdist\gap^{-1}$ pairwise distance calculations
to partition the $\gap^{-1}+1$ segments into $\numdist$ groups.  
By Remark~1, $\hat{d}(\cdot,\cdot)$ 
has computational complexity of $\O(n\operatorname{polylog}n)$.
The remaining calculations are of order 
$\O(\numdist(\gap^{-1}+1))$. 
This brings the resource complexity of the proposed algorithm to $\O(n^2\operatorname{polylog}n)$.
\section{Proof of Theorem~\ref{thm1}}\label{sec:proof}
In this section we prove the consistency of the proposed algorithm. 
The proof relies on a Lemma~\ref{prop0}.
We introduce the following additional notation. 
Consider the set $\S$ of segments 
specified by \eqref{alg:main:eq:segs} 
in Algorithm~\ref{alg:main}. 
For every segment $\widetilde{\x}_i:=X_{\psi_{i-1}..\psi_i} \in \S$ where $i=1..\numkappa+1$  
define $\widetilde{\rho}_i$ as the process distribution that generates the largest portion of $\widetilde{x}_i$;
that is, first define 
\begin{equation*}
K:=\argmaxdisp{ k \in \G_{\numdist'}}|\{\psi_{i-1}+1,\dots,\psi_i\} \cap \{n\theta_{k-1}+1,\dots,n\theta_{k}\}|
\end{equation*}
and then let $\widetilde{\rho}_i:= \rho_j$ where $j$ is such that $K\in \G_j$,
and $\G_j,~j=1..r$ are the ground-truth partitions defined by \eqref{eq:gt}.
\begin{lem}\label{prop0}
Let $\x \in \X^n,~n\in\N$ be a sequence with $\kappa$ change-points at least $\mingap$ apart for some $\mingap \in (0,1)$. 
Assume that the distributions that generate $\x$ are stationary and ergodic. 
Let $\S$ be the set of segments specified by  \eqref{alg:main:eq:segs} in Algorithm~\ref{alg:main}.
For all $\gap \in (0,\mingap)$ with probability one we have
\begin{equation*}
\lim_{n\rightarrow \infty} \sup_{\substack{\x_i \in \S}}\hat{d}(\widetilde\x_i,\widetilde{\rho}_i) = 0
\end{equation*}
\end{lem}
\begin{proof}
Fix an $\epsilon \in (0,\gap/2)$. 
There exists some $T$ such that 
\begin{equation}\label{prop:eq:ml}
\sum_{m,l=T}^\infty w_m w_l \leq \epsilon. 
\end{equation}
Moreover, for every $n \geq T/\gap$ and $m \in 1..T$ we have 
\begin{equation}\label{prop:eq:mbound}
\frac{m}{n\gap}\leq \epsilon.
\end{equation}  
For simplicity of notation define $\pi_k:=n\theta_k,~k=1..\kappa$.
Since the initial set of change-point candidates 
are produced by a {\em consistent} list-generator $\Upsilon$ (see Definition~\ref{defn:gen}), 
there exists an index-set 
$\I:=\{\mu_1,\dots,\mu_{\kappa}\} \in \{1..\numkappa\}^\kappa$ and some $N_0$ such that for all $n \geq N_0$ we have
\begin{equation}\label{prop0:eq:constmuk}
\sup_{k=1..\kappa}\frac{1}{n}|\psi_{\mu_k}-\pi_k| \leq \epsilon.
\end{equation}
Moreover, the initial candidates 
are at least $n\gap$ apart so that
\begin{equation}\label{prop0:eq:linlen}
\inf_{i\in1..\numkappa+1}\psi_i-\psi_{i-1}\geq n\gap
\end{equation}
where $\psi_0:=0$ and  $\psi_{\numkappa+1}:=n$. 
Let $\I':=\{1..\numkappa\}\setminus \I$. By \eqref{prop0:eq:constmuk} and \eqref{prop0:eq:linlen} for all $n \geq N_0$ 
the candidates indexed by $\I'$ have linear distances 
from the true change-points. 
\begin{align}\label{prop:eq:mingap}
\inf_{\substack{k\in1..\kappa\\ i\in \I'}}|\pi_k-\psi_i| \geq 
\inf_{\substack{k\in1..\kappa\\i\in \I',j\in \I}}|\psi_i-\psi_j|-|\pi_k-\psi_j| 
\geq n (\gap-\epsilon)
\end{align}
Denote by $\S_1:=\{\widetilde{x}_i:=X_{\psi_{i-1}+1..\psi_{i}} \in \S: \{i,i-1\}\cap \I = \emptyset\}$ 
the subset of the segments in $\S$ whose elements are formed by joining pairs of {\em consecutive elements }
of $\I'$ and let $\S_2:=\S\setminus \S_1$ be its complement. 
Let the true change-points that 
appear immediately to the left and to the right of 
an index $j \in 1..n-1$ be given by
\begin{align*} 
\L(j):=\max_{k\in 0..\kappa+1} \pi_k \leq j ~\text{and}~\RR(j):=\min_{k\in 0..\kappa+1}\pi_k > j
\end{align*} 
respectively, with $\pi_0:=0,~\pi_{\kappa+1}:=n$ where equality occurs when $j$ is itself a change-point. 
\textbf{1.~}Consider $\widetilde{x}_i:=X_{\psi_{i-1}+1..\psi_i} \in \S_1$. 
Observe that by definition $\widetilde{x}_i$ cannot contain a true change-point for $n\geq N_0$
since otherwise either $i-1$ or $i$ would belong to $\I$ 
contradicting the assumption that $\widetilde{x}_i\in\S_1$. 
Therefore for all $n \geq N_0$ we have $\widetilde{\rho}_i=\rho$ where 
$\rho \in \{\rho_1,\dots,\rho_{\numdist}\}$ is the process distribution that generates $X_{\L(\psi_{i-1})..\RR({\psi}_{i-1})}$.
To show that $\hat{d}(\widetilde{x}_i,\rho)\leq\epsilon$ we proceed as follows.
For each $m,l \in 1..\N$ 
we can find a finite subset $\beta^{m,l}$ of $B^{m,l}$ such that 
$\rho(\beta^{m,l})\geq 1-\epsilon$.
Observe that the segments $X_{\L(\psi_{i-1})..b}$ have lengths 
at least $\gap n$ for all $b\in \L(\psi_{i-1})+n\gap..\RR(\psi_{i-1})$.  
Therefore, for every $B \in \beta^{m,l},~m,l \in \N$ there exists some $N(B)$ 
such that for all $n\geq N(B)$ with probability $1$ we have 
\begin{equation}\label{prop0:eq:freq}
\sup_{b\in \L(\psi_{i-1})+n\gap..\RR(\psi_{i-1})}|\nu(X_{\L(\psi_{i-1})..b},B)-\rho(B)|\leq \epsilon.
\end{equation}
Using the definition of $\nu(\cdot,\cdot)$ given by \eqref{eq:nu} we obtain the
following algebraic manipulation of 
the frequency function. 
For every $B \in B^{m,l},~m,l \in \N$ 
 we have
 \begin{small}
 \begin{align}\label{eq:numan}
\nu(\widetilde{x}_i,B)=& 
\frac{\psi_i-\L(\psi_{i-1})-m+1}{\psi_i-\psi_{i-1}-m+1}\nu(X_{\L(\psi_{i-1})+1..\psi_{i}},B) \\
&-\frac{\psi_{i-1}-\L(\psi_{i-1})-m+1}{\psi_i-\psi_{i-1}-m+1}
 \nu(X_{\L(\psi_{i-1})+1..\psi_{i-1}},B) 
 -\hspace{-0.5cm}\sum_{j=\psi_{i-1}-m+1}^{\psi_{i-1}} 
\hspace{-0.1cm}\frac{\mathbb{I}\{X_{j..j+m} \in B\}}{\psi_i-\psi_{i-1}-m+1}\notag
 \end{align}
 \end{small}
 where the last summation is upper bounded (in absolute value) by $ \frac{m-1}{\psi_{i}-\psi_{i-1}-m+1}$. 
Let $N'_i:=\max\{N_0,\maxdisp{B\in \beta^{m,l},~m,l\in1..T}N(B),\frac{T}{\epsilon \gap}\}$. 
For all $n \geq N_i'$ we have
\begin{align}
\hat{d}(\widetilde{x}_i,\rho) &=\sum_{m,l=1}^{\infty}w_{m,l}\sum_{B \in B^{m,l}}|\nu(\widetilde{x}_i,B)-\rho(B)|\notag\\
&\leq\sum_{m,l=1}^{T}w_{m,l}\hspace{-0.3cm}\sum_{B \in \beta^{m,l}}\hspace{-0.2cm}\frac{\psi_{i}-\psi_{i-1}-m+1}{\psi_{i}-\psi_{i-1}}|\nu(\widetilde{x}_i,B)-\rho(B)|+\frac{m-1}{\psi_{i}-\psi_{i-1}}+2\epsilon \label{p2}\\ 
&\leq\sum_{m,l=1}^{T}w_{m,l}\hspace{-0.3cm}\sum_{B \in \beta^{m,l}}\hspace{-0.2cm}
\frac{\psi_i-\L(\psi_{i-1})-m+1}{\psi_i-\psi_{i-1}}|\nu(X_{\L(\psi_{i-1})+1..\psi_i},B)-\rho(B)| \label{p3}\\
&\quad+\frac{\psi_{i-1}-\L(\psi_{i-1})-m+1}{\psi_{i}-\psi_{i-1}}
 |\nu(X_{\L(\psi_{i-1})+1..\psi_{i-1}},B)-\rho(B)|+\frac{2(m-1)}{\psi_{i}-\psi_{i-1}}+2\epsilon \notag\\
&\leq2\epsilon(2+\gap^{-1})\label{p4}
\end{align}
where \eqref{p2} follows from \eqref{prop:eq:ml}, the definition of $\beta^{m,l}$ and the fact that $|\nu(\cdot,\cdot)-\rho(\cdot)|\leq 1$; 
\eqref{p3} follows from \eqref{eq:numan}, 
and \eqref{p4} follows from \eqref{prop:eq:mbound}, \eqref{prop0:eq:linlen}, and \eqref{prop0:eq:freq}. \\
Let $N':=\max_{i\in|\S_1|}N_i'$. For all $n\geq N'$ we have
\begin{equation}\label{prop:firsthalf}
\sup_{\widetilde{x}_i\in \S_1}\hat{d}(\widetilde{x}_i,\widetilde{\rho}_i) \leq 2\epsilon(2+\gap^{-1}) .
\end{equation}
\textbf{2.~}Take $\widetilde{x}_i:=X_{\psi_{i-1}..\psi_i} \in \S_2$. 
Observe that by definition $\I\cap\{i,i-1\}\neq \emptyset$ 
so that either $i-1$ or $i$ belong to $\I$.
We prove the statement for the case where 
$i-1\in \I$. The case where $i \in \I$ is analogous. 
We start by showing that $[\psi_{i-1},\psi_i] \subseteq [\pi-\epsilon,\pi'+\epsilon]$
for all $n \geq N_0$ where, 
\begin{equation*}
\pi:=\argmindisp{\pi_{k},k=1..\kappa}\frac{1}{n}|\pi_{k}-\psi_{i-1}|~\text{and}~\pi':=\RR(\pi).
\end{equation*}
Since $i-1 \in \I$, by \eqref{prop0:eq:constmuk} for all $n\geq N_0$ we have 
$\frac{1}{n}|\pi-\psi_{i-1}| \leq \epsilon$.
We have two cases. 
Either $i \in \I$ so that  by \eqref{prop0:eq:constmuk}
for all $n \geq N_0$ we have $\frac{1}{n}|\psi_i-\pi'|\leq \epsilon$,
or $i \in \I'$ in which case $\psi_i < \pi'$.  
To see the latter statement assume by way of contradiction that $\psi_i > \pi'$
where $\pi' \neq n$; (the statement trivially holds for $\pi'=n$). 
By the consistency of $\Upsilon$ there exists some $j > {i-1} \in \I$
such that $\frac{1}{n}|\psi_j-\pi'|\leq \epsilon$ for all $n \geq N_0$. 
Thus from \eqref{prop0:eq:constmuk} and \eqref{prop:eq:mingap} 
we obtain that $\psi_i-\psi_j \geq \gap-2\epsilon >0$. 
Since the initial estimates are sorted in increasing order, 
this implies $j \leq i$ leading to a contradiction. 
Thus we have $[\psi_{i-1},\psi_i] \subseteq [\pi-\epsilon,\pi'+\epsilon]$
so that $\widetilde{\rho}_i=\rho$ where $\rho$ is the process distribution 
$\rho \in \{\rho_1,\dots,\rho_\numdist \}$ that generates $X_{\pi..\pi'}$. 
To show that $\hat{d}(\widetilde{x}_i,\rho)\leq \epsilon$ we proceed as follows. 
Let $\pi'':=\min\{\psi_i,\pi'\}$. 
It is easy to see that by \eqref{defn:thetamin}, \eqref{prop:eq:mingap}, 
and the assumptions that $\mingap >0$ and $\gap \in (0,\mingap)$
the segment $X_{\pi..\pi''}$ has length at least $n\gap$. Therefore, 
for each $m,l \in 1..\N$ 
we can find a finite subset $\beta^{m,l}$ of $B^{m,l}$ such that 
$\rho(\beta^{m,l})\geq 1-\epsilon$. For every $B \in \beta^{m,l},~m,l \in \N$ there exists some $N'(B)$ 
such that for all $n\geq N'(B)$ we have 
\begin{equation}\label{prop:eq:freq}
|\nu(X_{\pi+1..\pi''},B)-\rho(B)|\leq \epsilon.
\end{equation}
For every $B \in B^{m,l},~m,l \in 1..T$ 
we have the following algebraic manipulation of $\nu(\widetilde{x}_i,B)$.
\begin{small}
\begin{align}\label{prop:newdec}
&\nu(\widetilde{x}_i,B)=\frac{\pi''-\pi-m+1}{\psi_i-\psi_{i-1}-m+1}
\nu(X_{\pi+1..\pi''},B)
+\frac{\mathbb{I}\{\psi_i >\pi'\}}{\psi_i-\psi_{i-1}-m+1} \sum_{j=\pi'+1}^{\psi_i-m+1} \mathbb{I}\{X_{j..j+m}\in B\} \\
&+\frac{\mathbb{I}\{\psi_{i-1} <\pi\}}{\psi_i-\psi_{i-1}-m+1} \sum_{j= \psi_{i-1}+1 }^{ \pi -m+1} \mathbb{I}\{X_{j..j+m}\in B\}
-\frac{\mathbb{I}\{\psi_{i-1} >\pi\}}{\psi_i-\psi_{i-1}-m+1} \hspace{-0.5cm}\sum_{j=\pi+1}^{ \psi_{i-1} -m+1} \hspace{-0.4cm}\mathbb{I}\{X_{j..j+m}\in B\}\notag
\end{align}
\end{small}\\
For all 
$B \in \beta^{m,l},~m,l \in 1..T$ and all $n \geq \max\{N_0,\max_{B \in \beta^{m,l},~m,l \in 1..T}N'(B)\}$ we have,
\begin{small}
\begin{align}\label{prop:numanbound}
\frac{\psi_i-\psi_{i-1}-m+1}{\psi_i-\psi_{i-1}}|\nu(\widetilde{x}_i,B)-\rho(B)| 
\leq &\frac{\pi''-\pi-m+1}{\psi_i-\psi_{i-1}}|\nu(X_{\pi+1..\pi''},B)-\rho(B)|\notag \\
&+\frac{n(\psi_i-\pi'')}{\psi_i-\psi_{i-1}}+\frac{n|\psi_i-\pi|}{\psi_i-\psi_{i-1}}\leq3\epsilon\gap^{-1}
\end{align}
\end{small}
\noindent where the first inequality follows from  \eqref{prop:newdec} 
and the second inequality follows from \eqref{prop0:eq:constmuk}, \eqref{prop0:eq:linlen} and \eqref{prop:eq:freq}.
Let $N_i'':= \max\{N_0,\max_{B \in \beta^{m,l},~m,l \in 1..T}N'(B),\frac{T}{\epsilon\gap}\}$. For all $n\geq N_i''$ we have,
\begin{align}
\hat{d}(\widetilde{x}_i,\rho) 
&\leq\sum_{m,l=1}^{T}w_{m,l} \sum_{B \in \beta^{m,l}}
\frac{\psi_i-\psi_{i-1}-m+1}{\psi_i-\psi_{i-1}}|\nu(\widetilde{\x}_i,B)-\rho(B)| +\frac{m-1}{\psi_i-\psi_{i-1}}+2\epsilon \label{prop:3}\\
&\leq 2\epsilon(1+2\gap^{-1})\notag 
\end{align}
where the first inequality follows from \eqref{prop:eq:ml}, the definition of $\beta^{m,l}$ and observing that $|\nu(\cdot,\cdot)-\rho(\cdot)|\leq 1$
and the second inequality follows from\eqref{prop:eq:mbound}, \eqref{prop0:eq:linlen} and \eqref{prop:numanbound}.\\
Let $N'':=\max_{i: \x_i \in \S_2(\epsilon)}N_i''$. 
For $n\geq N''$ we have
\begin{equation}\label{prop:secondhalf}
\sup_{\widetilde{\x}_i\in\S_2} \hat{d}(\widetilde{\x}_i,\widetilde{\rho}_i)  \leq 2\epsilon(1+2\gap^{-1}).
\end{equation}
Finally, by \eqref{prop:firsthalf} and \eqref{prop:secondhalf} for all $n\geq \max\{N',N''\}$ 
we have 
$\sup_{\widetilde{\x}_i\in\S } \hat{d}(\widetilde{\x}_i,\widetilde{\rho}_i)  \leq 2\epsilon(3+2\gap^{-1})$
Since  $\epsilon$  can be chosen  arbitrary small, this proves the statement.
\end{proof}
\begin{proof} (of Theorem~\ref{thm1})
Let $\delta:=\min_{r'\neq r'' \in 1..\numdist}d(\rho_{r'},\rho_{r''})$ denote the minimum distance
between the distinct distributions that generate $\x$.
Fix an $\epsilon\in(0,\delta/4)$. By  Lemma~\ref{prop0} and applying the triangle inequality 
there exists some $N_1$ such that for all $n \geq N_1$ we have
\begin{align}
\inf_{\substack{\widetilde{\x}_i,\widetilde{\x}_j \in \S,~\widetilde{\rho}_i\neq\widetilde{\rho}_j }}\hat{d}(\widetilde{\x}_i,\widetilde{\x}_j)\geq  \delta -2\epsilon, \text{~and~} 
\sup_{\substack{\widetilde{\x}_i,\widetilde{\x}_j \in \S,~\widetilde{\rho}_i=\widetilde{\rho}_j }}\hat{d}(\widetilde{\x}_i,\widetilde{\x}_j)\leq 2\epsilon. \label{thm:same}
\end{align}
Let $\pi_k:=n\theta_k,~k=1.\kappa$. 
By the consistency of $\Upsilon$ (see Definition~\ref{defn:gen} and Proposition~\ref{thm0}) there exists some $N_2$ 
such that for all $n \geq N_2$ there exists a set 
$\{\mu_1,\dots,\mu_\kappa\} \in \{1..\kappa\}^{\numkappa}$
such that 
\begin{equation}\label{thm_eq_const}
\frac{1}{n}|\psi_{\mu_k}-\pi_k| \leq \epsilon.
\end{equation}
Let $N:= \max N_i,~i=1,2$. 
By \eqref{thm:same} for all $n \geq N$ we have  
\begin{equation}
\widetilde{\rho}_{c_j} \neq \widetilde{\rho}_{c_{j'}},~j=2..\numdist,~j'=1..j-1
\end{equation}
where $c_j,~j=1..\numdist$ is given by \eqref{alg:cj}.
Hence, the cluster centers $\x_{c_j}, j=1..r$  
 are each generated by a different process distribution.
On the other hand, 
the rest of the segments are each assigned to the closest cluster,  
so that  
from~(\ref{thm:same}) for all $n \geq N$ we have 
\begin{equation}\label{thm:const_tag}
T(\widetilde{\x}_i)=T(\widetilde{\x}_{i'}) \Leftrightarrow \widetilde{\rho}_i=\widetilde{\rho}_{i'}.
\end{equation}
By construction the index-set $\mathcal C$ generated
by Algorithm~\ref{alg:main} corresponds to  {\em those and only those} change-point 
candidates that separate consecutive segments assigned to {\em different} 
clusters, by \eqref{thm:const_tag} 
for all $n \geq N$ and all $i \in \mathcal C$ we have 
$\widetilde{\rho}_i \neq \widetilde{\rho}_{i+1}$.
Thus $\hat{\kappa}=\kappa$ and $\hat{\theta}_k=\frac{1}{n}\psi_{\mu_k},~k=1..\kappa$. 
Notice that 
by~\eqref{thm_eq_const} $\psi_{\mu_k},~k=1..\kappa$ are  consistent
estimates of $\pi_k$.
\end{proof}
\section{Experimental Results}\label{sec:exp}
In this section we present empirical evaluations of our
algorithms on synthetically generated data. 
To generate the data we use stationary ergodic process distributions that do not belong to any ``simpler" 
general class of time-series, and cannot be approximated by finite state models. 
In particular they cannot be modeled by hidden Markov process distributions with finite state-spaces. 
Moreover,  the single-dimensional
marginals of all distributions are the same throughout the generated sequence. 
Similar  distribution families  are  commonly used as examples in this framework  \citep[see, e.g.,][]{Sheilds:96}.  
\begin{figure}[!ht]\label{fig:synth}
\begin{center}
\includegraphics[width=0.7\textwidth]{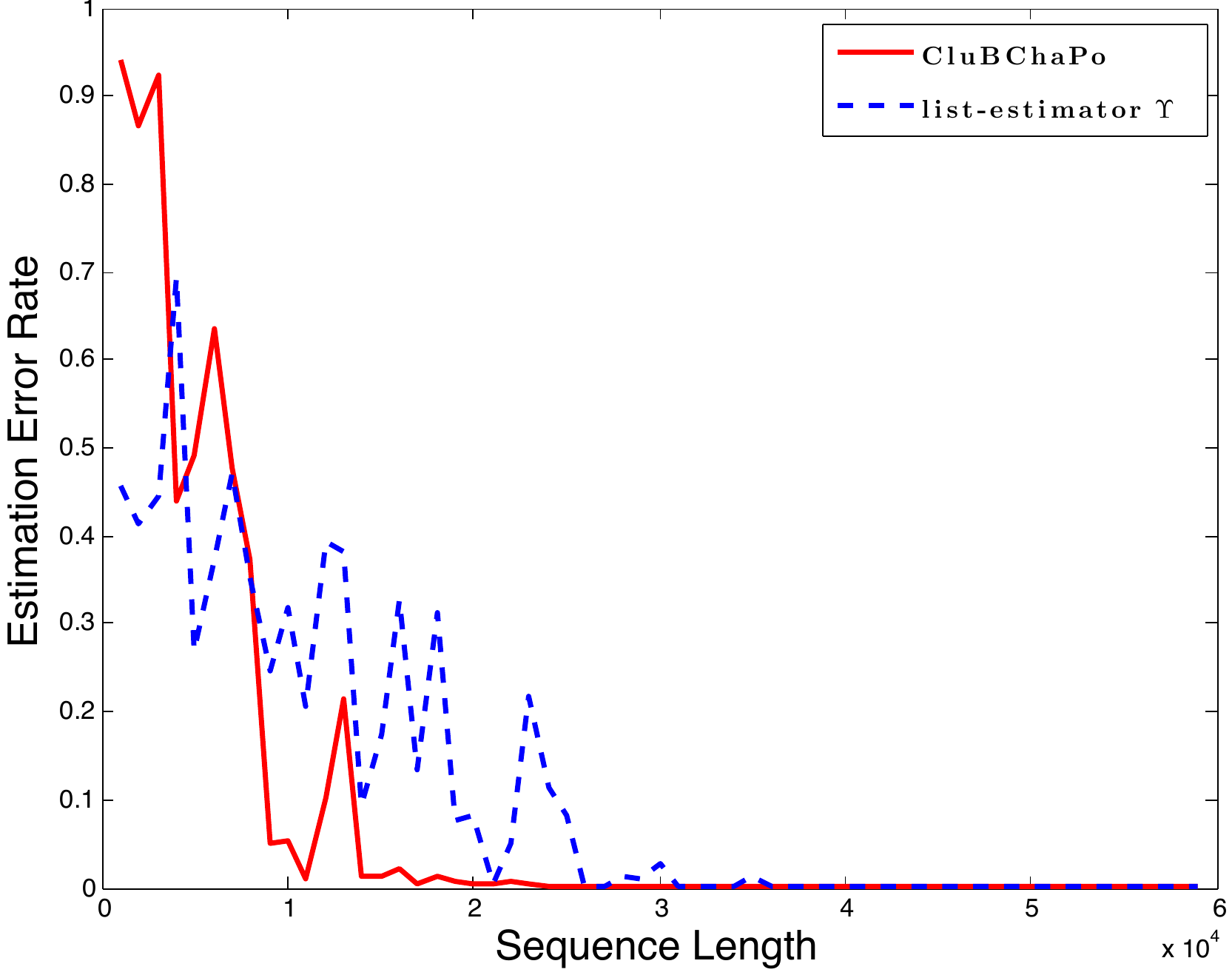}
\caption{{Average (over $40$ runs) error rates of our algorithm CluBChaPo$(\x,\gap,r)$ 
and the list-estimator $\Upsilon$ of \cite{khaleghi:12mce}, 
as a function of the length $n$ of the input sequence $\x \in \R^n$, where
$\x$ has $\kappa=4$ change-points $\mingap:=0.1$ apart and is generated by $r=3$ distributions; 
$\gap:=0.6 \mingap$.
The error of $\Upsilon(\x,\gap)$ is based on its first $\kappa$ elements.}}
\end{center}
\label{fig1}
\end{figure}
The distributions and the  procedure to generate a sequence $\x:=X_1,\dots,X_m \in \R^m,~m\in \N$  are as follows. 
Fix a parameter $\alpha \in (0,1)$ and two uniform distributions $\U_1$ and $\U_2$. 
Let $r_0$ be drawn randomly from $[0,1]$.
For each $i=1..m$ obtain $r_i:=r_{i-1}+\alpha \mod 1$ 
and draw $x^{(j)}_i$ from $\U_j,~j=1,2$. Finally set 
$X_i:=\mathbb{I}\{r_i\leq 0.5\}x_i^{(1)}+\mathbb{I}\{r_i> 0.5\}x_i^{(2)}$.
If $\alpha$ is irrational\footnote{$\alpha$ is simulated by a long double with a long mantissa.
} this produces a real-valued stationary ergodic time-series.
In the  experiments we fixed three parameters 
$\alpha_1:=0.12..,~\alpha_2:=0.13..$ and $\alpha_3:=0.14..$ { (with long mantissae)}
to correspond to $r=3$ different process distributions. 
To produce $\x \in \R^n$ we randomly generated 
$\kappa:=5$ change-points $\theta_k,~k=1..\kappa$ at least $\mingap$ apart, with  $\mingap:=0.1$.
Every segment of length $n_k:=n(\theta_k-\theta_{k-1}),~k=1..\kappa+1$ with $\theta_0:=0,~\theta_{\kappa+1}:=1$ 
was generated with $\alpha_{k'}$ and $n_k$ where $k':=k\mod r,~k=0..\kappa+1$.
In our experiments we provide $\gap:=0.6\mingap$ as input 
and calculate the error as 
$\mathbb{I}\{|\mathcal C|\neq\kappa\}+\mathbb{I}\{|{\mathcal C|=\kappa\}}\sum_{k=1}^{\kappa}|\hat{\theta}_{k}-\theta_k|$. 

Note that, with this data generation procedure,
the single-dimensional marginals are the same throughout 
the sequence. Most of the existing algorithms 
do not work at all in this scenario. To the best of our knowledge,
the only work to address the change-point problem under this general 
framework is that of \cite{khaleghi:12mce}, which we use here for comparison.
However, this method is a list-estimator in the sense of Definition~\ref{defn:gen}
and makes no attempt to estimate $\kappa$. 
It simply generates a sorted list of estimates, 
whose first $\kappa$ elements converge to the true change-points;
we calculate the error on the first $\kappa$ elements of its output.

\section{Discussion}\label{sec:conc}
We have presented an asymptotically consistent 
method to  estimate the number of change-points 
and do  locate the changes in highly dependent time-series data.
The considered framework is very general and as such is suitable for real-world applications. 

Note that in this setting rates of convergence (even of frequencies to respective probabilities) 
are provably impossible 
to obtain. Therefore, unlike in the traditional settings for change-point analysis, 
the algorithms developed for  this framework are forced not to rely on any rates of convergence. 
We see this as an 
advantage of the framework as it means that the algorithms are applicable to a 
much wider range of situations.
At the same time, it may be interesting to derive the rates of convergence of the proposed algorithm 
under stronger assumptions (e.g.,  i.i.d.\  data, or some mixing conditions). 
We conjecture that the  algorithm is indeed optimal (up to some constant factors) in such settings
as well (although it clearly cannot be optimal under parametric assumptions); however, we leave this as future work.

In the proposed algorithm a specific consistent clustering method is used to estimate the 
number of change-points. 
An interesting extension would be to establish the consistency of this method 
using any list-estimator in combination with 
any time-series clustering algorithm, 
that possess suitable asymptotic  consistency guarantees. 

Finally, the consistency of the algorithm is established when the distributional distance is 
used as the distance between the segments. The proof relies on some properties specific to this distance. 
Other distances can also be used in problems concerning stationary ergodic time series \citep[e.g.,][]{Ryabko:12red};
thus, it is interesting to investigate which distances can be used with the algorithm proposed in the current paper.

\end{document}